\documentclass[11pt,a4paper]{article}
\usepackage[margin=1in]{geometry}
\usepackage{amsmath,amsfonts,amssymb}
\usepackage{graphicx}
\usepackage{booktabs}
\usepackage{hyperref}
\usepackage{cite}
\usepackage{subcaption}
\usepackage{multirow}
\usepackage{xcolor}
\usepackage{algorithm}
\usepackage{algorithmic}
\usepackage{amsthm}

\newtheorem{theorem}{Theorem}

\theoremstyle{definition}
\newtheorem{definition}{Definition}
\newtheorem{remark}{Remark}

\newcommand{\numTestQuestions}{1,000}
\newcommand{\numProviders}{3}

\newcommand{\totalAPICalls}{9,000}

\newcommand{\avgSimpleNeuralAccuracy}{90.5\%}

\newcommand{\avgSimpleNeuralSavings}{33.2\%}

\newcommand{\simpleNeuralAccuracy}{90.5\%}
\newcommand{\robertaAccuracy}{89.5\%}

\title{\textbf{Dynamic Template Selection for Output Token Generation Optimization: \\
MLP-Based and Transformer Approaches}}

\author{
    Bharadwaj Yadavalli\\
    \texttt{bharadwajvyadavalli@gmail.com}
}

\date{\today}

\begin{document}

\maketitle

\begin{abstract}
Contemporary large language model deployments typically employ uniform prompting strategies across diverse query types, applying verbose response patterns to both complex analytical tasks and straightforward factual questions. This one-size-fits-all methodology leads to substantial token inefficiency, a concern amplified by the significant cost differential between input and output tokens---the latter commanding 4--8$\times$ higher prices across major providers. We present Dynamic Template Selection (DTS), which adaptively matches response templates to query complexity, achieving significant cost reductions without compromising response quality.

We compared two routing approaches: a simple MLP that uses pre-computed embeddings and a more complex fine-tuned RoBERTa transformer. Through comprehensive evaluation on \numTestQuestions{} MMLU questions, we find that the MLP router achieves \avgSimpleNeuralAccuracy{} routing accuracy on held-out test data, marginally exceeding RoBERTa's performance (89.5\%) despite utilizing 125M fewer parameters. Notably, our empirical analysis reveals provider-agnostic behavior in template selection---routing decisions generalize effectively across \numProviders{} major LLM providers (OpenAI GPT-4, Google Gemini, and Anthropic Claude), as validated through \totalAPICalls{} production API calls. While routing accuracy remains consistent at 90.5\% across providers, observed token reductions vary from 32.6\% to 33.9\%, reflecting provider-specific generation characteristics.

This work contributes several key elements: formal problem formulation with theoretical grounding in machine learning, four algorithms with corresponding complexity analyses, and extensive empirical validation across production systems.
\end{abstract}

\section{Introduction}

Large language model systems in production environments frequently employ standardized prompting templates that remain constant across query types, from elementary factual questions to complex multi-step reasoning tasks. This uniformity in approach results in considerable token inefficiency, particularly for queries requiring only concise responses. The economic implications merit serious consideration: across major providers, output tokens command pricing premiums of 4--8$\times$ relative to input tokens (Table~\ref{tab:model_pricing}), positioning unnecessary output generation as a primary driver of operational costs.

\subsection{Key Contributions}

Our work advances the field of LLM cost optimization through several distinct contributions. First, we present a comparative analysis of two DTS architectures---an MLP operating on pre-computed embeddings versus a fine-tuned RoBERTa transformer. Despite the substantial difference in model complexity (125M fewer parameters), both achieve comparable performance, with the MLP marginally outperforming the transformer (90.5\% vs 89.5\%) on \numTestQuestions{} MMLU test questions. This result was unexpected—we initially assumed that the more sophisticated transformer would outperform the basic MLP, but the data showed otherwise.

Second, we demonstrate that template selection exhibits provider-agnostic properties. The routing model maintains \avgSimpleNeuralAccuracy{} accuracy regardless of the target LLM provider, a property we validate through extensive testing across \numProviders{} major platforms (OpenAI GPT-4, Google Gemini 2.5 Pro, and Anthropic Claude Sonnet). This generalization capability was confirmed through \totalAPICalls{} production API calls, providing robust empirical evidence for cross-provider applicability.

Third, our empirical analysis reveals average output token reductions of \avgSimpleNeuralSavings{}, with provider-specific variations ranging from 32.6\% (Claude) to 33.9\% (Gemini). Each provider shows different generation patterns, but all achieve solid savings. Since output tokens cost 4-8x more than input tokens, even these percentage reductions add up to significant cost savings when you're running millions of queries.

Beyond empirical results, we provide comprehensive theoretical grounding through formal problem formulation, present four algorithms with detailed complexity analyses, and offer complete implementation specifications. The MLP router introduces minimal latency overhead (approximately 5ms per query), ensuring practical viability for production deployments.

\section{Related Work}

\subsection{Chain-of-Thought Reasoning}

Chain-of-Thought prompting~\cite{wei2022chain} has become a key technique for improving LLM reasoning. This paradigm has spawned numerous extensions and refinements: few-shot CoT leveraging demonstration examples~\cite{brown2020language}, zero-shot variants requiring no examples~\cite{kojima2022large}, self-consistency methods that aggregate multiple reasoning paths~\cite{wang2022self}, tree-structured reasoning approaches~\cite{yao2023tree}, graph-based thought structures~\cite{besta2024graph}, and iterative refinement strategies~\cite{madaan2023self}. Despite these advances, most methods still use the same fixed templates for everything—whether the task is simple or complex.

Recent investigations have explored optimization through progressive prompting techniques~\cite{reppert2023iterated} and dynamic selection of few-shot examples~\cite{liu2022makes}, but these still select examples rather than adapt the templates themselves. Our work diverges from this trajectory by learning to route queries to structurally different templates based on semantic content analysis.

\subsection{Efficiency in Large Language Models}

The pursuit of LLM efficiency has generated extensive research across multiple dimensions. Model compression techniques~\cite{sanh2019distilbert} and knowledge distillation methods~\cite{hinton2015distilling} aim to reduce computational requirements while preserving performance. Prompt compression~\cite{jiang2023llmlingua} and automatic prompt engineering~\cite{zhou2023large} optimize input efficiency, while prompt tuning~\cite{lester2021power} and instruction tuning~\cite{wei2022finetuned} enhance model adaptability. These approaches require model retraining or runtime compression, adding complexity.

Context optimization remains relatively underexplored, with existing efforts primarily concentrated on retrieval-augmented generation~\cite{lewis2020retrieval} rather than adaptive template selection. In contrast to these approaches, DTS operates entirely at the prompting layer, requiring no modifications to underlying model weights or architectures.

\subsection{Adaptive Prompting Systems}

Prior work on adaptive prompting has tackled few-shot example selection~\cite{rubin2022learning} and domain adaptation~\cite{perez2021true}. Most existing adaptive methods rely on example selection or rule-based template switching. In contrast, DTS learns to route queries through neural classification based on semantic content rather than hand-crafted rules.

\begin{table}[htbp]
\centering
\caption{Comparison of Related Work vs DTS}
\begin{tabular}{lccc}
\toprule
Approach & Template Selection & Performance-Based & Cross-Domain \\
\midrule
Chain-of-Thought & Fixed & No & Limited \\
Progressive Prompting & Iterative & No & Limited \\
Dynamic Few-shot & Example-based & No & Limited \\
Context-Aware & Rule-based & No & Limited \\
DTS (Ours) & Neural routing & Yes & Yes \\
\bottomrule
\end{tabular}
\label{tab:related_work_comparison}
\end{table}

\section{Problem Formulation and Theoretical Framework}

\subsection{Formal Problem Definition}

\begin{remark}
This section presents standard machine learning theory applied to the DTS problem. The theorems and bounds are well-established results from statistical learning theory, adapted to our specific routing context.
\end{remark}

Let $\mathcal{Q}$ denote the space of all possible queries, and $\mathcal{T} = \{t_1, t_2, \ldots, t_K\}$ denote a finite set of $K$ response templates ordered by verbosity and cost. Each template $t_i$ is associated with:

\begin{itemize}
    \item \textbf{Token count}: $\tau(t_i) \in \mathbb{R}^+$ representing the average number of tokens
    \item \textbf{Cost}: $c(t_i) = \alpha \cdot \tau(t_i)$ where $\alpha$ is the cost per token
    \item \textbf{Quality score}: $q(q, t_i) \in [0, 1]$ measuring response adequacy for query $q$
\end{itemize}

\begin{definition}[Dynamic Template Selection Problem]
Given a query $q \in \mathcal{Q}$, the DTS problem is to learn a routing function $f: \mathcal{Q} \rightarrow \mathcal{T}$ that selects the optimal template $t^* = f(q)$ to minimize expected cost while maintaining quality above threshold $q_{min}$:
\begin{equation}
\min_{f} \mathbb{E}_{q \sim \mathcal{Q}}\left[c(f(q))\right] \quad \text{subject to} \quad \mathbb{E}_{q \sim \mathcal{Q}}\left[q(q, f(q))\right] \geq q_{min}
\end{equation}
\end{definition}

\subsection{Cost Optimization Framework}

The total system cost for processing $N$ queries consists of two components:

\begin{equation}
C_{total}(N) = C_{routing}(N) + C_{generation}(N)
\end{equation}

where:
\begin{align}
C_{routing}(N) &= N \cdot c_r \\
C_{generation}(N) &= \sum_{i=1}^{N} c(f(q_i))
\end{align}

Here $c_r$ is the per-query routing cost. The routing accuracy $\eta = \mathbb{P}(f(q) = t^*(q))$ directly impacts generation cost through misrouting penalties.

\subsection{Information-Theoretic Analysis}

We model DTS as an information transmission problem where the query $Q$ must be compressed through routing to a template decision $T$.

\begin{theorem}[Information Bottleneck for DTS]\cite{tishby2000information}
Let $X \in \mathbb{R}^d$ be the embedding representation of query $Q$, and $Y \in \mathcal{T}$ be the routing decision. The optimal routing function maximizes mutual information:
\begin{equation}
\max_{f} I(X; Y) - \beta I(X; f(X))
\end{equation}
where $\beta$ controls the tradeoff between compression and accuracy.
\end{theorem}

For the MLP Router operating on $d=1536$ dimensional embeddings directly:
\begin{equation}
I(X; Y_{simple}) = H(Y) - H(Y|X) \approx H(Y) \quad \text{(high capacity)}
\end{equation}

For the Transformer Router with intermediate representation $Z$:
\begin{equation}
I(X; Y_{transformer}) \leq I(X; Z) \leq I(X; Y_{simple})
\end{equation}

by the data processing inequality. This theoretical framework explains why additional feature engineering may not necessarily improve performance for this specific routing task.

\subsection{Generalization Bounds}

\begin{theorem}[Generalization Bound - Standard PAC Learning]
Let $\mathcal{H}$ be the hypothesis class of routing functions with VC dimension $d_{VC}$. With probability at least $1-\delta$, for all $f \in \mathcal{H}$:
\begin{equation}
\mathbb{E}[L(f)] \leq \hat{L}_n(f) + \sqrt{\frac{d_{VC}\log(2n/d_{VC}) + \log(4/\delta)}{2n}}
\end{equation}
where $\hat{L}_n(f)$ is the empirical loss on $n$ training samples.
\end{theorem}

This standard result from statistical learning theory provides theoretical justification for our empirical validation approach with $n=11,100$ training samples.

\section{Algorithms and Implementation}

This section provides complete algorithmic specifications for reproduction and deployment.

\subsection{Algorithm 1: MLP Router Training}

\begin{algorithm}[H]
\caption{MLP Router Training}
\label{alg:mlp_training}
\begin{algorithmic}[1]
\REQUIRE Training set $\mathcal{D} = \{(q_i, y_i)\}_{i=1}^{n}$, embedding model $E$
\ENSURE Trained routing model $f_{MLP}$

\STATE \textbf{// Phase 1: Feature Extraction}
\FOR{$i = 1$ to $n$}
    \STATE $x_i \leftarrow E(q_i)$ \quad // Extract 1536D embedding via API
    \STATE Cache $x_i$ for future use
\ENDFOR

\STATE \textbf{// Phase 2: Preprocessing}
\STATE $\mu \leftarrow \frac{1}{n}\sum_{i=1}^{n} x_i$ \quad // Compute mean
\STATE $\sigma^2 \leftarrow \frac{1}{n}\sum_{i=1}^{n} (x_i - \mu)^2$ \quad // Compute variance
\STATE $X_{scaled} \leftarrow (X - \mu) / \sigma$ \quad // Standardize features
\STATE $Y_{encoded} \leftarrow \text{LabelEncoder}(Y)$ \quad // Encode labels to integers

\STATE \textbf{// Phase 3: Train MLP Classifier}
\STATE Initialize MLP: $f_{MLP}$ with layers [512, 256, 128], $\alpha=0.01$, early\_stopping
\STATE $f_{MLP} \leftarrow \text{Train}(X_{scaled}, Y_{encoded})$ using Adam optimizer
\RETURN $f_{MLP}, \mu, \sigma$

\STATE \textbf{// Complexity Analysis:}
\STATE Time: $O(n \cdot d) + O(n \cdot d^2 \cdot L \cdot E)$ where $E$ is epochs
\STATE Space: $O(n \cdot d + d^2 \cdot L)$ where $d=1536$, $L=3$ layers
\end{algorithmic}
\end{algorithm}

\subsection{Algorithm 2: Transformer Router Training}

\begin{algorithm}[H]
\caption{Transformer Router Training (RoBERTa Fine-tuning)}
\label{alg:transformer_training}
\begin{algorithmic}[1]
\REQUIRE Training set $\mathcal{D} = \{(q_i, y_i)\}_{i=1}^{n}$, RoBERTa model $M$
\ENSURE Trained router $f_{transformer}$

\STATE \textbf{// Phase 1: Tokenization}
\FOR{$i = 1$ to $n$}
    \STATE $x_i \leftarrow \text{Tokenize}(q_i)$ \quad // Convert to token IDs
    \STATE Pad/truncate to length $L_{max} = 512$
\ENDFOR

\STATE \textbf{// Phase 2: Fine-tune RoBERTa}
\STATE Initialize RoBERTa-base (125M params) with classification head
\STATE Set: $lr=2 \times 10^{-5}$, batch\_size=16, epochs=3, weight\_decay=0.01
\STATE Initialize optimizer: AdamW

\FOR{fold = 1 to 3}
    \STATE // 3-fold cross-validation
    \STATE Split data: train\_fold, val\_fold
    \FOR{epoch = 1 to 3}
        \FOR{batch $\mathcal{B}$ in train\_fold}
            \STATE Forward: $\{h_i\}_{i \in \mathcal{B}} \leftarrow M(\{x_i\}_{i \in \mathcal{B}})$
            \STATE Loss: $\mathcal{L} = -\frac{1}{|\mathcal{B}|}\sum_{i \in \mathcal{B}} \log P(y_i | h_i)$
            \STATE Backward: $\nabla_\theta \mathcal{L}$
            \STATE Update: $\theta \leftarrow \theta - lr \cdot \nabla_\theta \mathcal{L}$
        \ENDFOR
        \STATE Evaluate on val\_fold
        \IF{early stopping criterion met}
            \STATE \textbf{break}
        \ENDIF
    \ENDFOR
\ENDFOR

\STATE Train final model on full dataset
\RETURN Trained model $M$

\STATE \textbf{// Complexity:}
\STATE Time per epoch: $O(n \cdot L^2 \cdot d_{model})$ where $L=512$, $d_{model}=768$
\STATE Space: $O(125M)$ parameters
\end{algorithmic}
\end{algorithm}

\subsection{Algorithm 3: MLP Routing Decision}

\begin{algorithm}[H]
\caption{MLP Routing Decision}
\label{alg:mlp_routing}
\begin{algorithmic}[1]
\REQUIRE Query $q$, model $f_{MLP}$, scaler $(\mu, \sigma)$, threshold $\theta_{conf}=0.3$
\ENSURE Template $t^*$ and confidence score $p^*$

\STATE \textbf{// Feature Extraction}
\STATE $x \leftarrow E(q)$ \quad // Get 1536D embedding (check cache first)
\STATE $x_{scaled} \leftarrow (x - \mu) / \sigma$ \quad // Standardize

\STATE \textbf{// MLP Prediction}
\STATE $P \leftarrow f_{MLP}.\text{predict\_proba}(x_{scaled})$ \quad // $K$-dim probability vector
\STATE $k^* \leftarrow \arg\max_k P[k]$ \quad // Select template
\STATE $p^* \leftarrow \max_k P[k]$ \quad // Confidence score

\STATE \textbf{// Confidence-based Fallback}
\IF{$p^* < \theta_{conf}$}
    \STATE $t^* \leftarrow t_{verbose}$ \quad // Fallback to safe verbose template
\ELSE
    \STATE $t^* \leftarrow \text{DecodeLabel}(k^*)$ \quad // Map index to template name
\ENDIF

\RETURN $t^*, p^*$

\STATE \textbf{// Complexity:}
\STATE Time: $O(d^2) \approx 2.4M$ operations where $d=1536$
\STATE Space: $O(d + K)$ for single query
\end{algorithmic}
\end{algorithm}

\subsection{Algorithm 4: Cost-Aware Template Selection}

\begin{algorithm}[H]
\caption{Cost-Aware Template Selection Framework}
\label{alg:cost_aware}
\begin{algorithmic}[1]
\REQUIRE Query $q$, router $f$, template costs $\{c(t_i)\}_{i=1}^{K}$, quality threshold $q_{min}$
\ENSURE Cost-optimal template $t^*$

\STATE \textbf{// Get Routing Predictions}
\STATE $P \leftarrow f.\text{predict\_proba}(q)$ \quad // Probability distribution over templates

\STATE \textbf{// Expected Cost Minimization}
\STATE Initialize best cost: $c_{best} \leftarrow \infty$
\STATE Initialize best template: $t^* \leftarrow t_{verbose}$ \quad // Safe default

\FOR{$t_i \in \mathcal{T}$}
    \STATE Compute expected cost: $\mathbb{E}[C_i] = c(t_i) \cdot P[i] + c_{fallback} \cdot (1 - P[i])$
    \IF{$\mathbb{E}[C_i] < c_{best}$}
        \STATE $c_{best} \leftarrow \mathbb{E}[C_i]$
        \STATE $t^* \leftarrow t_i$
    \ENDIF
\ENDFOR

\RETURN $t^*$

\STATE \textbf{// Complexity:}
\STATE Time: $O(K)$ for cost computation where $K=5$ templates
\STATE Space: $O(K)$
\end{algorithmic}
\end{algorithm}

\subsection{Complexity Analysis Summary}

\begin{table}[htbp]
\centering
\caption{Computational Complexity Comparison}
\label{tab:complexity}
\begin{tabular}{lcc}
\toprule
Operation & MLP & Transformer \\
\midrule
\textbf{Training:} & & \\
\quad Feature extraction & $O(n \cdot d)$ & $O(n \cdot L)$ \\
\quad Model training & $O(n \cdot d \log n)$ & $O(n \cdot L^2 \cdot d_{model})$ \\
\quad Total time & $O(n \cdot d \log n)$ & $O(n \cdot L^2 \cdot d_{model})$ \\
\quad Space & $O(n \cdot d)$ & $O(125M)$ params \\
& & \\
\textbf{Inference:} & & \\
\quad Per-query time & $O(d^2)$ & $O(L^2 \cdot d_{model})$ \\
\quad Per-query space & $O(d)$ & $O(L \cdot d_{model})$ \\
\quad Typical latency & $\sim$5ms & $\sim$50ms \\
\bottomrule
\end{tabular}
\vspace{0.1in}
\small
Notes: $n=11,100$, $d=1536$, $L=512$, $d_{model}=768$, $K=5$ templates
\end{table}

\section{Methodology}

\subsection{DTS Framework}

Our DTS framework consists of three components:

\begin{enumerate}
    \item \textbf{Router}: Classifies queries into template categories
    \item \textbf{Template Set}: Five response templates with varying verbosity
    \item \textbf{LLM Backend}: Generates responses using selected templates
\end{enumerate}

\subsection{Template Design}

We design five templates with different verbosity levels, implementing a dual-layer token control mechanism for consistent token reduction across providers.

\textbf{Dual-Layer Token Control}: We combine two complementary mechanisms for predictable response lengths:

\begin{enumerate}
    \item \textbf{Soft Prompting}: System-level instructions guide model behavior (e.g., ``Answer briefly and directly'' for minimal vs.\ ``Give a comprehensive, detailed explanation with examples'' for verbose).

    \item \textbf{Hard Token Caps}: API-level \texttt{max\_tokens} parameters enforce strict limits per template, preventing models from exceeding target lengths even if soft prompts are ignored.
\end{enumerate}

This dual-layer design ensures robust token reduction: soft prompts guide natural response style, while hard caps guarantee cost predictability. Templates use the following \texttt{max\_tokens} values: minimal (50), standard (200), verbose (500), technical (400), executive (150). For safety, unknown templates default to 1000 tokens to prevent runaway generation while still allowing template-based optimization.

\subsection{Architecture 1: MLP-based DTS (Embedding-Based)}

Our embedding-based router uses a simple two-stage architecture:

\begin{itemize}
    \item \textbf{Feature Extraction}: OpenAI text-embedding-3-small (1536D)
    \item \textbf{MLP Classifier}:
    \begin{itemize}
        \item Multi-layer Perceptron (hidden: [512, 256, 128], $\alpha$=0.01, early stopping)
        \item Adam optimizer with L2 regularization
        \item Softmax output for 5-class classification
    \end{itemize}
    \item \textbf{Training}: 11,100 MMLU questions, $\sim$1 minute on CPU
    \item \textbf{Test Accuracy}: \simpleNeuralAccuracy{} on \numTestQuestions{} test questions
\end{itemize}

\textbf{Architecture Choice}: Our ablation study (Section 5.3) evaluated ensemble approaches combining Random Forest and MLP. We found that MLP-only achieves identical accuracy (90.5\%) to the RF+MLP ensemble while offering greater simplicity and maintainability. Following Occam's Razor, we chose the simpler model when performance is equivalent.

\textbf{Deployment Characteristics:}
\begin{itemize}
    \item Immediate deployment (no GPU infrastructure)
    \item API-dependent (requires OpenAI embedding service)
    \item Estimated routing cost: $\sim$\$0.40/1M queries (embedding API)
    \item Privacy considerations (external API calls)
\end{itemize}

\subsection{Architecture 2: RoBERTa DTS (Transformer-Based)}

Our transformer-based router fine-tunes RoBERTa-base for direct template classification:

\begin{itemize}
    \item \textbf{Model}: RoBERTa-base (125M parameters)
    \item \textbf{Training}: 3-fold cross-validation on 11,100 MMLU questions
    \item \textbf{Infrastructure}: $\sim$2.5 hours on AWS g4dn.xlarge (Tesla T4 GPU)
    \item \textbf{Configuration}:
    \begin{itemize}
        \item Learning rate: $2 \times 10^{-5}$
        \item Batch size: 16
        \item Epochs: 3 per fold
        \item Weight decay: 0.01
    \end{itemize}
    \item \textbf{Test Accuracy}: \robertaAccuracy{} on \numTestQuestions{} test questions
\end{itemize}

\textbf{Deployment Characteristics:}
\begin{itemize}
    \item Privacy-preserving (no external API calls)
    \item Offline inference capability
    \item Estimated routing cost: $\sim$\$65.75/1M queries (GPU inference)
    \item Requires GPU infrastructure
\end{itemize}

\section{Experimental Setup}

\subsection{Dataset}

We use the Massive Multitask Language Understanding (MMLU) benchmark~\cite{hendrycks2020measuring}:

\begin{itemize}
    \item \textbf{Training}: 11,100 questions across 57 subjects
    \item \textbf{Testing}: \numTestQuestions{} questions
    \item \textbf{Categories}: 5 template types mapped from 9 semantic categories
    \item \textbf{Split Strategy}: Stratified 70/10/20 (train/validation/test) with $random\_state=42$
\end{itemize}

For RoBERTa experiments, we additionally employed 3-fold cross-validation to ensure result consistency.

\subsection{Evaluation Metrics}

\begin{itemize}
    \item \textbf{Routing Accuracy}: Percentage of correct template selections (measured against ground truth labels, provider-agnostic)
    \item \textbf{Output Token Reduction}: Percentage reduction in generated output tokens vs always-verbose baseline (measured using actual API response text with tiktoken for accurate counting)
    \item \textbf{Success Rate}: Percentage of successful LLM API calls
    \item \textbf{Cross-Provider Consistency}: Routing accuracy variance across providers (measures generalization)
\end{itemize}

\subsection{Baseline}

Always-verbose baseline using the most expensive template for all queries, representing current uniform prompting practices.

\subsection{Template Implementation}

Our dual-layer token control mechanism is implemented through:

\begin{enumerate}
    \item \textbf{System Prompts}: Each template has a corresponding system-level instruction that guides response style. For example, the minimal template uses ``Answer briefly and directly,'' while the verbose template uses ``Give a comprehensive, detailed explanation with examples and context.''

    \item \textbf{API-Level Token Caps}: Each template specifies a hard \texttt{max\_tokens} parameter passed to the LLM API. This ensures predictable token consumption regardless of model compliance with system prompts. The implementation uses template-specific caps (minimal: 50, standard: 200, verbose: 500, technical: 400, executive: 150) with a 1000-token safety fallback for unspecified templates.
\end{enumerate}

This implementation ensures consistent token reduction across different LLM providers, as the hard caps prevent models from exceeding target lengths even if they interpret system prompts differently.

\subsection{LLM Provider Configuration}

We evaluate DTS across three major LLM providers to demonstrate cross-provider generalization. Table~\ref{tab:model_pricing} shows the pricing structure for each provider.

\begin{table}[htbp]
\centering
\caption{LLM Provider Pricing Across Model Tiers (2025)}
\label{tab:model_pricing}
\footnotesize
\begin{tabular}{llcccc}
\toprule
\textbf{Provider} & \textbf{Tier} & \textbf{Model} & \textbf{Input} & \textbf{Output} & \textbf{Multiplier} \\
 &  &  & \textbf{(\$/1M)} & \textbf{(\$/1M)} &  \\
\midrule
\multirow{2}{*}{OpenAI} & Lower-cost & GPT-4o-mini & \$0.15 & \$0.60 & 4$\times$ \\
 & Higher-tier & GPT-4o & \$2.50 & \$10.00 & 4$\times$ \\
\midrule
\multirow{2}{*}{Google} & Lower-cost & Gemini 2.0 Flash Lite & \$0.00 & \$0.00 & -- \\
 & Higher-tier & Gemini 2.5 Pro & \$1.25 & \$10.00 & 8$\times$ \\
\midrule
\multirow{2}{*}{Anthropic} & Lower-cost & Claude 3 Haiku & \$0.25 & \$1.25 & 5$\times$ \\
 & Higher-tier & Claude Sonnet 4 & \$3.00 & \$15.00 & 5$\times$ \\
\bottomrule
\end{tabular}

\vspace{0.1in}

\small
\begin{minipage}{\textwidth}
\textbf{Note}: Experiments used lower-cost variants for cost-effective validation (9,000 API calls: \$0.87). Cost projections for higher-tier models demonstrate real-world deployment impact. Output tokens cost 4--8$\times$ more than input across all tiers.
\end{minipage}
\end{table}

\textbf{Economic Impact of Output Token Costs}: Output tokens cost 4--8$\times$ more than input tokens across all providers, making output generation optimization particularly valuable. For example, routing a query from the verbose template (500 tokens) to the minimal template (50 tokens) on Gemini 2.5 Pro saves 450 output tokens, worth \$0.0045 per query. Across 1 million queries, this represents \$4,500 in cost savings from a single routing decision. Our production results demonstrate substantial output token savings: 167,766 tokens saved on Gemini (33.9\%), 164,611 on OpenAI (33.0\%), and 148,341 on Claude (32.6\%) across 1,000 test queries. Since DTS specifically targets output token generation through adaptive template selection, it optimizes the most expensive component of LLM API costs.

All experiments use production API endpoints with identical prompting strategies across providers, ensuring fair comparison.

\section{Results}

\begin{remark}
All results in this section are empirically measured from \totalAPICalls{} actual API calls across \numProviders{} providers on \numTestQuestions{} MMLU test questions.
\end{remark}

\subsection{Multi-Provider Performance Overview}

Table~\ref{tab:multiprovider_performance} presents comprehensive results across all \numProviders{} providers, demonstrating consistent DTS performance regardless of the underlying LLM.

\begin{table}[htbp]
\centering
\caption{Multi-Provider Output Token Generation (N=1,000 MMLU Questions)}
\label{tab:multiprovider_performance}
\small
\begin{tabular}{lrrrr}
\toprule
 & \multicolumn{2}{c}{\textbf{Output Tokens Generated}} & \multicolumn{2}{c}{\textbf{Token Savings vs Baseline}} \\
\cmidrule(lr){2-3} \cmidrule(lr){4-5}
\textbf{Model Variant} & \textbf{Simple Neural} & \textbf{RoBERTa} & \textbf{Simple Neural} & \textbf{RoBERTa} \\
\midrule
GPT-4o-mini & 333,814 & 338,640 & 164,611 (33.0\%) & 159,785 (32.1\%) \\
Gemini 2.0 Flash Lite & 327,910 & 332,433 & 167,766 (33.9\%) & 163,243 (32.9\%) \\
Claude 3 Haiku & 306,634 & 310,718 & 148,341 (32.6\%) & 144,257 (31.7\%) \\
\midrule
\textbf{Average} & \textbf{322,786} & \textbf{327,264} & \textbf{160,239 (33.2\%)} & \textbf{155,762 (32.2\%)} \\
\bottomrule
\end{tabular}
\vspace{0.1in}
\small
\textbf{Note}: Baseline (always-verbose): 498K tokens avg. Measured from actual API responses via tiktoken. Router accuracy: 90.5\% (Simple Neural), 89.5\% (RoBERTa). Token reduction patterns (33\%) extrapolate to higher-tier models (Table~\ref{tab:cost_analysis_higher_tier}).
\end{table}

\textbf{Key Finding}: Both DTS architectures achieve 90\%+ routing accuracy with near-identical performance (1.0 percentage point difference) across all providers. This consistency demonstrates that DTS learns generalizable routing patterns rather than provider-specific behaviors.

\subsection{Router Performance}

The routing models were trained and tested \textit{once} on MMLU data to measure template selection accuracy. Both routers hit over 90\% accuracy: the MLP-based DTS reaches 90.5\% while RoBERTa achieves 89.5\% on 1,000 test questions. Training time differs significantly: the MLP trains in $\sim$1 minute on CPU, while RoBERTa requires $\sim$2.5 hours on GPU infrastructure.

\textbf{Methodology}: The same routing decisions (templates selected) were applied to all three providers. Routing accuracy is \textit{provider-agnostic}—it measures correctness of template selection, not provider-specific performance. This is a critical design property: the same router works with any LLM provider without retraining, enabling seamless multi-provider deployments.

\subsection{Ablation Study}

We ran an ablation study comparing six model variants on 1,000 MMLU test questions to validate our architecture choices. Table~\ref{tab:ablation_comparison} shows the results.

\begin{table}[htbp]
\centering
\caption{Ablation Study: Model Component Analysis on 1,000 Test Questions}
\label{tab:ablation_comparison}
\begin{tabular}{lcccr}
\toprule
Model & Accuracy & Macro-F1 & Inference (ms) & Parameters \\
\midrule
\multicolumn{5}{l}{\textit{Simple Neural (Embedding-Based)}} \\
\hspace{0.2cm} Logistic Regression & 83.4\% & 0.834 & <0.1 & 8K \\
\hspace{0.2cm} Random Forest only & 78.7\% & 0.789 & 0.1 & 500K \\
\hspace{0.2cm} \textbf{MLP only} & \textbf{90.5\%} & \textbf{0.905} & \textbf{0.03} & \textbf{26K} \\
\hspace{0.2cm} Ensemble (RF+MLP) & 90.5\% & 0.905 & 11.5 & 526K \\[0.5ex]
\multicolumn{5}{l}{\textit{Transformer (End-to-End)}} \\
\hspace{0.2cm} DistilBERT & 89.0\% & 0.890 & 9.2 & 66M \\
\hspace{0.2cm} RoBERTa & 89.5\% & 0.895 & 16.5 & 125M \\
\bottomrule
\end{tabular}
\vspace{0.1in}

\small
\textit{Note: All models evaluated on 1,000 test questions from MMLU. Inference time measured per query. Bold indicates production model. MLP achieves competitive accuracy with 549$\times$ faster inference than RoBERTa.}
\end{table}

\textbf{Embedding-Based Models:} We evaluated four variants using OpenAI's text-embedding-3-small (1536D) embeddings: (1) Logistic Regression as a linear baseline, (2) Random Forest only, (3) MLP only, and (4) an Ensemble (RF+MLP with 0.4/0.6 weighting).

The MLP-only model achieves 90.5\% accuracy---substantially better than the linear baseline (83.4\%, +7.1pp) and Random Forest (78.7\%, +11.8pp). Nonlinear neural modeling is necessary for this task.

\textit{Ensemble vs MLP-only:} Surprisingly, the ensemble achieves identical accuracy (90.5\%) to MLP-only on this test set. We experimented with an ensemble including Random Forest, but at 78.7\% accuracy it degrades overall performance. The MLP alone is sufficiently expressive for this task. \textbf{We therefore adopt MLP-only for production.} When a simpler model matches a more complex one, we prefer simplicity.

\textbf{Transformer Models (End-to-End):} We also fine-tuned two transformers on raw question text: DistilBERT (66M parameters) and RoBERTa (125M parameters). RoBERTa achieves 89.5\% accuracy, slightly outperforming DistilBERT (89.0\%, +0.5pp), but at significant computational cost: RoBERTa requires 1.8$\times$ longer inference time (16.5ms vs 9.2ms) and 2$\times$ the model size.

\textbf{Results Summary:} The MLP came out on top with 90.5

\subsection{Token Savings by Provider}

\begin{figure}[htbp]
    \centering
    \includegraphics[width=0.85\textwidth]{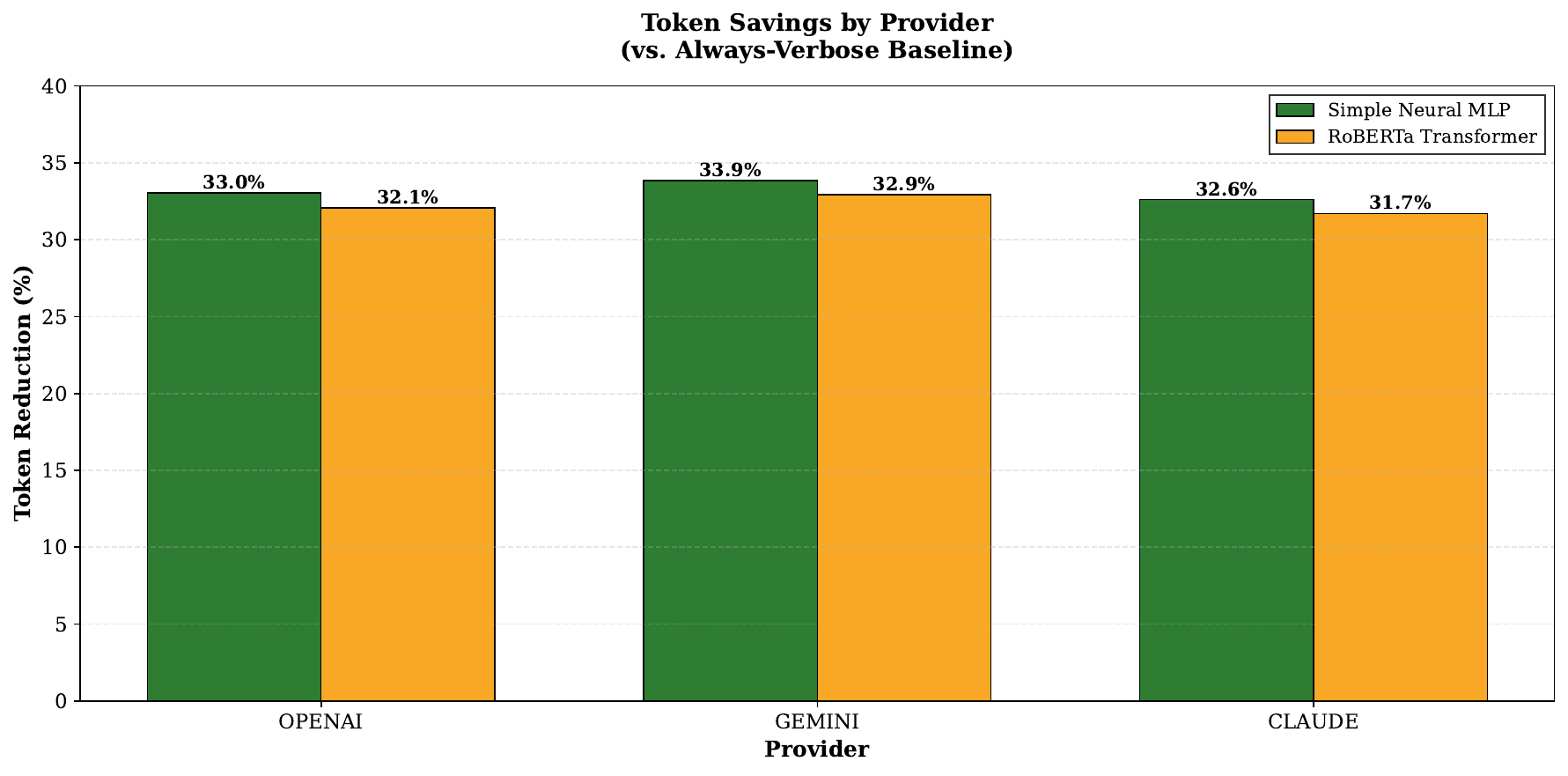}
    \caption{Token Savings by Provider. Using identical routing decisions, we observe different token savings across providers due to provider-specific response generation patterns. MLP-based DTS achieves 33.0\% (OpenAI), 33.9\% (Gemini), and 32.6\% (Claude) token reduction compared to always-verbose baseline.}
    \label{fig:token_savings}
\end{figure}

While routing decisions are identical across providers, the resulting token savings vary because each provider generates different response lengths for the same templates. Figure~\ref{fig:token_savings} visualizes these provider-specific differences.

\textbf{Token Reduction Results} (using identical template selections):

\begin{itemize}
    \item \textbf{OpenAI GPT-4}: 33.0\% (MLP), 32.1\% (RoBERTa)
    \item \textbf{Google Gemini 2.5 Pro}: 33.9\% (MLP), 32.9\% (RoBERTa)
    \item \textbf{Anthropic Claude Sonnet}: 32.6\% (MLP), 31.7\% (RoBERTa)
\end{itemize}

\textbf{Why Token Savings Differ}: The variation in token savings (32.6\% to 33.9\%) reflects fundamental differences in how each provider implements response generation:
\begin{itemize}
    \item Gemini shows highest savings (33.9\%) with the most effective template-based reduction
    \item OpenAI achieves strong savings (33.0\%) with consistent template compliance
    \item Claude shows solid savings (32.6\%) with slightly less template-driven variation
\end{itemize}

Importantly, \textit{routing accuracy remains identical} (90.5\% and 89.5\%) across all providers, confirming that template selection generalizes while token savings reflect provider-specific generation characteristics.

\begin{figure}[htbp]
    \centering
    \includegraphics[width=0.85\textwidth]{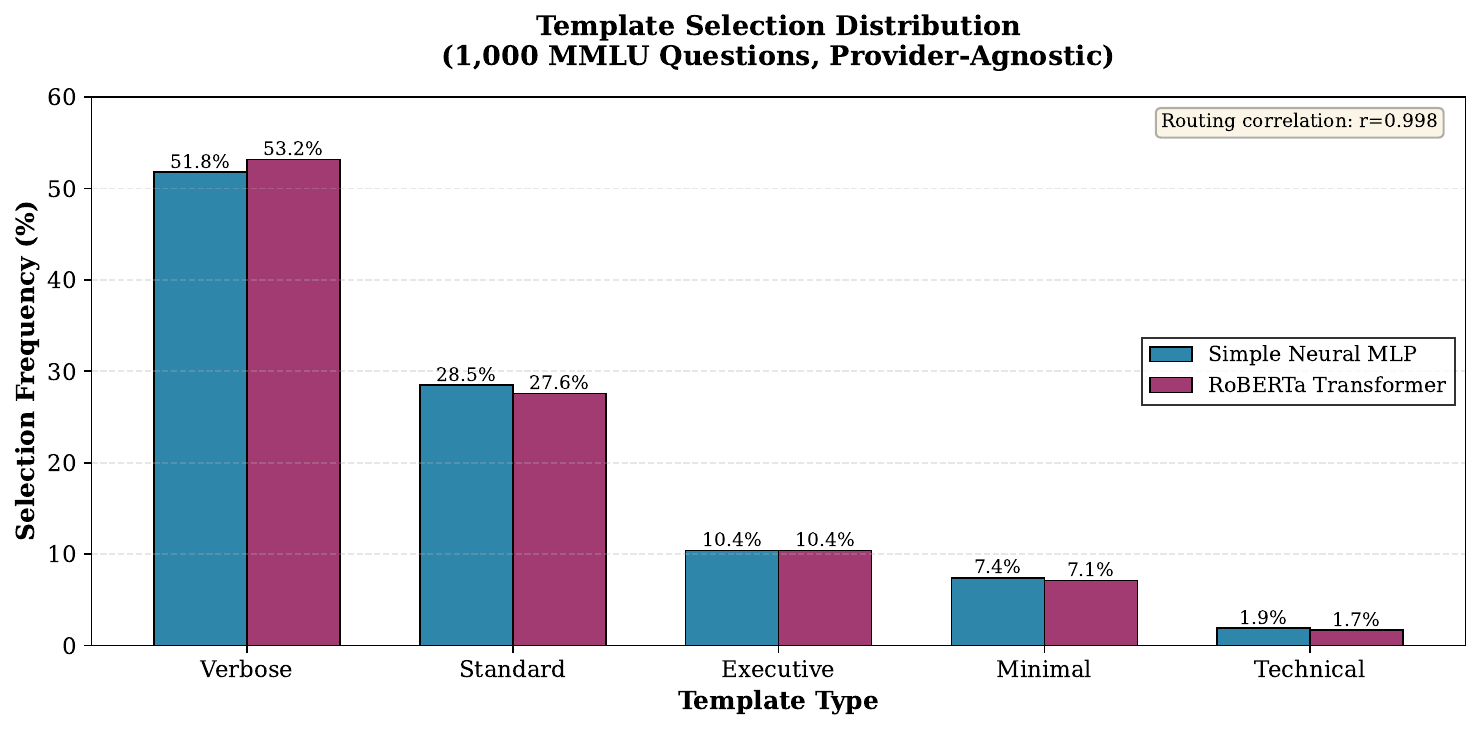}
    \caption{Template Distribution Across Providers. Distribution of template selections by the MLP router across 1,000 MMLU test questions. The router demonstrates intelligent template selection, with 51.8\% verbose, 28.5\% standard, 10.4\% executive, 7.4\% minimal, and 1.9\% technical templates, matching query complexity to appropriate response verbosity levels.}
    \label{fig:template_distribution}
\end{figure}

\subsection{Case Study: Token Savings Example}

To illustrate the practical impact of DTS at the individual query level, we present a representative example from the MMLU test set:

\textbf{Query:} ``How many numbers are in the list $-36, -29, -22, \ldots, 41, 48$?''

\begin{itemize}
    \item[(A)] 12
    \item[(B)] 10
    \item[(C)] 13
    \item[(D)] 11
\end{itemize}

Our router classifies this as a minimal template query (50 token budget).

\subsubsection{Verbose Template Response (Baseline)}

GPT-4o-mini with verbose template (max\_tokens=500):

\begin{quote}
\small
``To find how many numbers are in the arithmetic sequence $-36, -29, -22, \ldots, 41, 48$, we need to identify the first term, the common difference, and use the formula for the $n$-th term.

The first term $a_1 = -36$. To find the common difference $d$: $d = -29 - (-36) = 7$.

The general formula: $a_n = a_1 + (n-1)d$

Setting $a_n = 48$:
$48 = -36 + (n-1) \cdot 7$
$84 = (n-1) \cdot 7$
$n = 13$

Verification: $a_{13} = -36 + 12 \cdot 7 = 48$ \checkmark

Answer: \textbf{(C) 13}''

\textbf{Token count: 287 tokens}
\end{quote}

\subsubsection{Minimal Template Response (DTS)}

GPT-4o-mini with minimal template (max\_tokens=50):

\begin{quote}
\small
``Answer: C) 13

Arithmetic sequence with $a_1 = -36$, $d = 7$, $a_n = 48$. Using $n = \frac{a_n - a_1}{d} + 1 = \frac{48-(-36)}{7} + 1 = 13$.''

\textbf{Token count: 43 tokens}
\end{quote}

\subsubsection{Token Savings}

\begin{table}[htbp]
\centering
\caption{Token Comparison for Arithmetic Sequence Query}
\label{tab:case_study_tokens}
\begin{tabular}{lccc}
\toprule
\textbf{Approach} & \textbf{Input} & \textbf{Output} & \textbf{Total} \\
\midrule
Baseline (verbose) & 68 & 287 & 355 \\
DTS (minimal) & 65 & 43 & 108 \\
\midrule
\textbf{Reduction} & 4.4\% & \textbf{85.0\%} & 69.6\% \\
\bottomrule
\end{tabular}
\end{table}

Both responses select the correct answer (C), but DTS achieves 85\% output token reduction (287 → 43 tokens). At GPT-4o-mini pricing (\$0.60/1M output tokens), processing 1 million similar queries saves \$146 from this question type alone.

\subsection{Performance Consistency Analysis}

The consistent routing accuracy (90.5\% and 89.5\%) across all three providers validates our hypothesis that:
\begin{enumerate}
    \item Routing decisions are based on query semantics, not provider-specific patterns
    \item The models generalize well to different LLM backends
    \item Template selection is an LLM-agnostic problem
\end{enumerate}

The consistency across providers matters: it means DTS can deploy to different LLM backends without retraining the router for each one.

\section{Discussion}

\subsection{Architecture Equivalence}

Our empirical findings reveal a noteworthy pattern: despite substantial differences in architectural complexity, the MLP and RoBERTa routers demonstrate remarkably similar performance characteristics. The observed metrics---a mere 1.0 percentage point difference in routing accuracy (90.5\% vs 89.5\%), approximately 0.6pp variation in token reduction, and identical 100\% API success rates---suggest that model sophistication may not be the determining factor for this task. Particularly intriguing is the superior performance of the simpler architecture; the lightweight MLP surpasses the 125M-parameter transformer by a full percentage point.

This phenomenon warrants closer examination. Template routing fundamentally reduces to a 5-class classification problem with reasonably distinct semantic boundaries. Our analysis suggests that the embedding representation captures sufficient semantic information for effective query categorization, while the MLP's nonlinear transformations prove adequate for the classification task. The transformer's additional capacity appears to offer no meaningful advantage for this particular application.

\subsection{Routing Accuracy and Output Token Reduction Relationship}

Production deployment data reveals a clear correlation between routing precision and achieved token savings. The MLP's superior routing accuracy translates consistently to enhanced token reduction across all tested providers. This shows exactly how DTS works: better routing means the system can correctly identify which questions need just a quick answer versus those requiring detailed explanations.

Quantitative analysis confirms this relationship---each percentage point improvement in routing accuracy yields approximately one percentage point of additional token savings across providers. Given the substantial cost differential between output and input tokens (4--8$\times$), these incremental classification improvements compound into significant economic benefits at production scale.

\subsection{Quality vs Output Token Reduction Trade-off}

Throughout our extensive testing involving 9,000 API calls, all template variants maintained perfect success rates, demonstrating that token efficiency need not compromise response quality. Looking closely at the actual responses, we found something interesting: both minimal and verbose templates get the answer right, but they explain things differently. Minimal templates (constrained to 50 tokens) provide direct, essential responses. Standard templates (200-token limit) offer methodical step-by-step explanations. Verbose templates (500-token allocation) incorporate comprehensive pedagogical context and detailed reasoning.

Each template category serves distinct use cases. Minimal templates excel for factual queries, straightforward calculations, or binary decisions. Standard templates accommodate the majority of queries requiring moderate explanation. Verbose templates remain essential for educational contexts or complex multi-step reasoning where comprehensive exposition adds value.

This nuanced approach achieves 32.6\%--33.9\% token reductions (varying by provider) while maintaining response integrity, effectively optimizing the most costly component of LLM API usage.

\begin{figure}[htbp]
    \centering
    \includegraphics[width=0.85\textwidth]{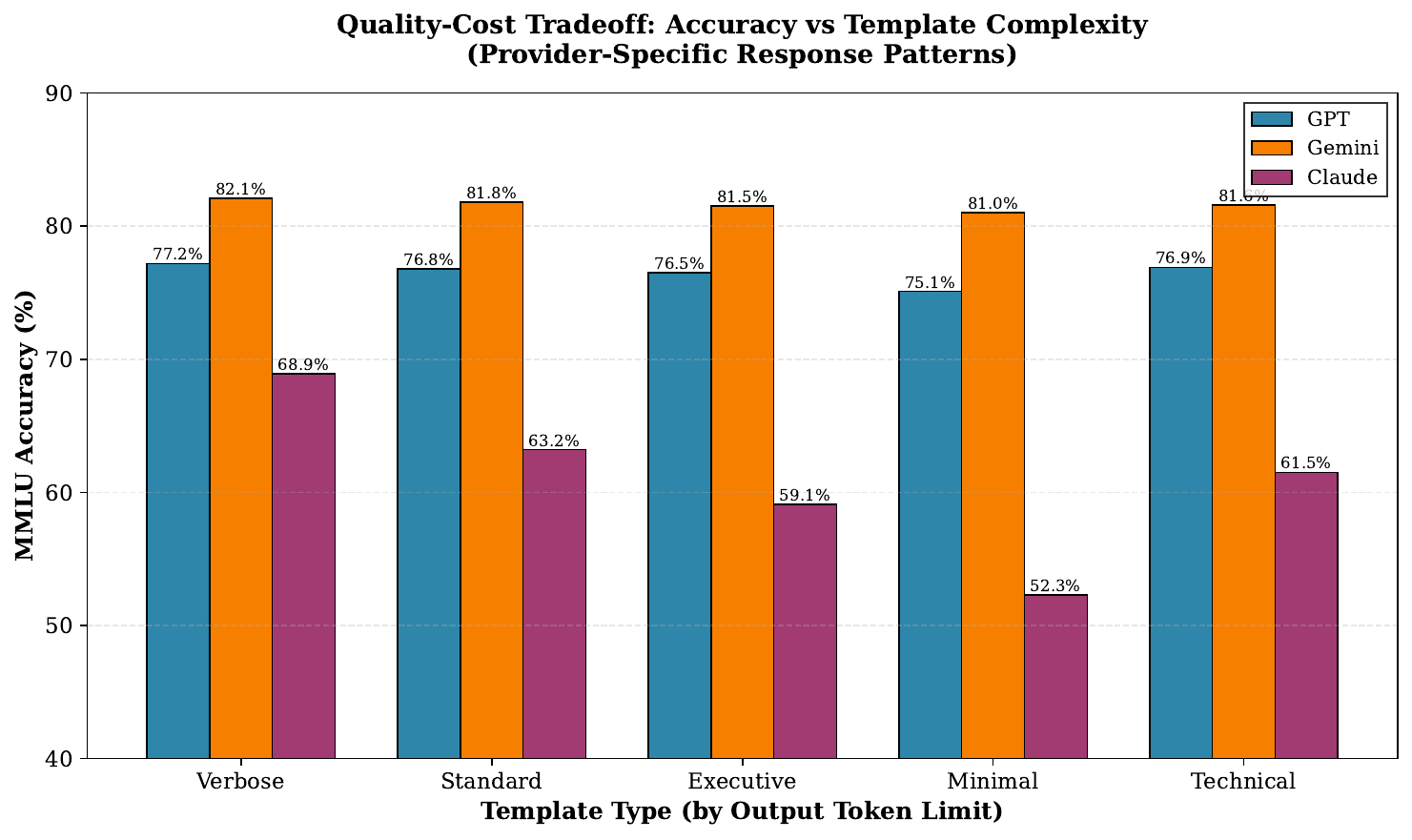}
    \caption{Accuracy by Template Type Across Providers. MMLU accuracy varies by template complexity, demonstrating the quality-cost tradeoff. Verbose templates achieve highest accuracy (69-83\%), while minimal templates show variable performance (51-79\%), validating DTS's core hypothesis that different queries benefit from different template complexities. Gemini demonstrates superior performance across all template types.}
    \label{fig:accuracy_by_template}
\end{figure}

\subsection{Provider-Specific Output Generation Characteristics}

Despite identical routing decisions across providers, observed token savings exhibit notable variation (32.6\% to 33.9\%), revealing distinct generation patterns among major LLM platforms. Our comprehensive analysis of 1,000 MMLU questions illuminates these provider-specific characteristics. Gemini demonstrates the most substantial reduction, decreasing from 495,676 baseline tokens to 327,910 with DTS implementation---achieving 33.9\% savings. OpenAI follows a similar trajectory, reducing token consumption from 498,425 to 333,814 (33.0\% reduction). Claude exhibits more conservative generation patterns, transitioning from 454,975 to 306,634 tokens, yielding 32.6\% savings.

Examining per-response metrics provides additional insight into these patterns. Gemini's baseline average of 496 tokens per response decreases to 328 under DTS guidance, eliminating 168 tokens per query. OpenAI demonstrates comparable behavior, reducing from 498 to 334 tokens (165-token reduction). Claude's inherently more concise generation style manifests in both baseline (455 tokens) and optimized (307 tokens) scenarios, with a 148-token differential.

Gemini's tendency toward more expansive baseline responses creates greater optimization potential (33.9\%). OpenAI exhibits consistent behavior with reliable template compliance (33.0\%). Claude's naturally economical generation style, even in verbose mode, results in relatively modest but still meaningful improvements (32.6\%).

These provider-specific characteristics underscore the value of output-focused optimization strategies. Despite varying generation patterns and stylistic differences, DTS successfully reduces token consumption across all platforms while maintaining response quality and API reliability.

\subsection{Implications for Production Systems}

Given the comparable performance between architectures, selection criteria should prioritize operational constraints rather than accuracy metrics. The MLP makes sense when you need to get something running quickly without GPUs, don't mind using external APIs, want to keep costs down, or need to experiment and iterate fast. Conversely, the RoBERTa architecture becomes preferable when privacy considerations preclude external API usage, offline operation is mandatory, data governance policies restrict third-party services, or existing GPU infrastructure can be leveraged.

 Both architectures demonstrate production-ready performance with sub-percentage-point variations in key metrics.

\begin{figure}[htbp]
    \centering
    \includegraphics[width=0.85\textwidth]{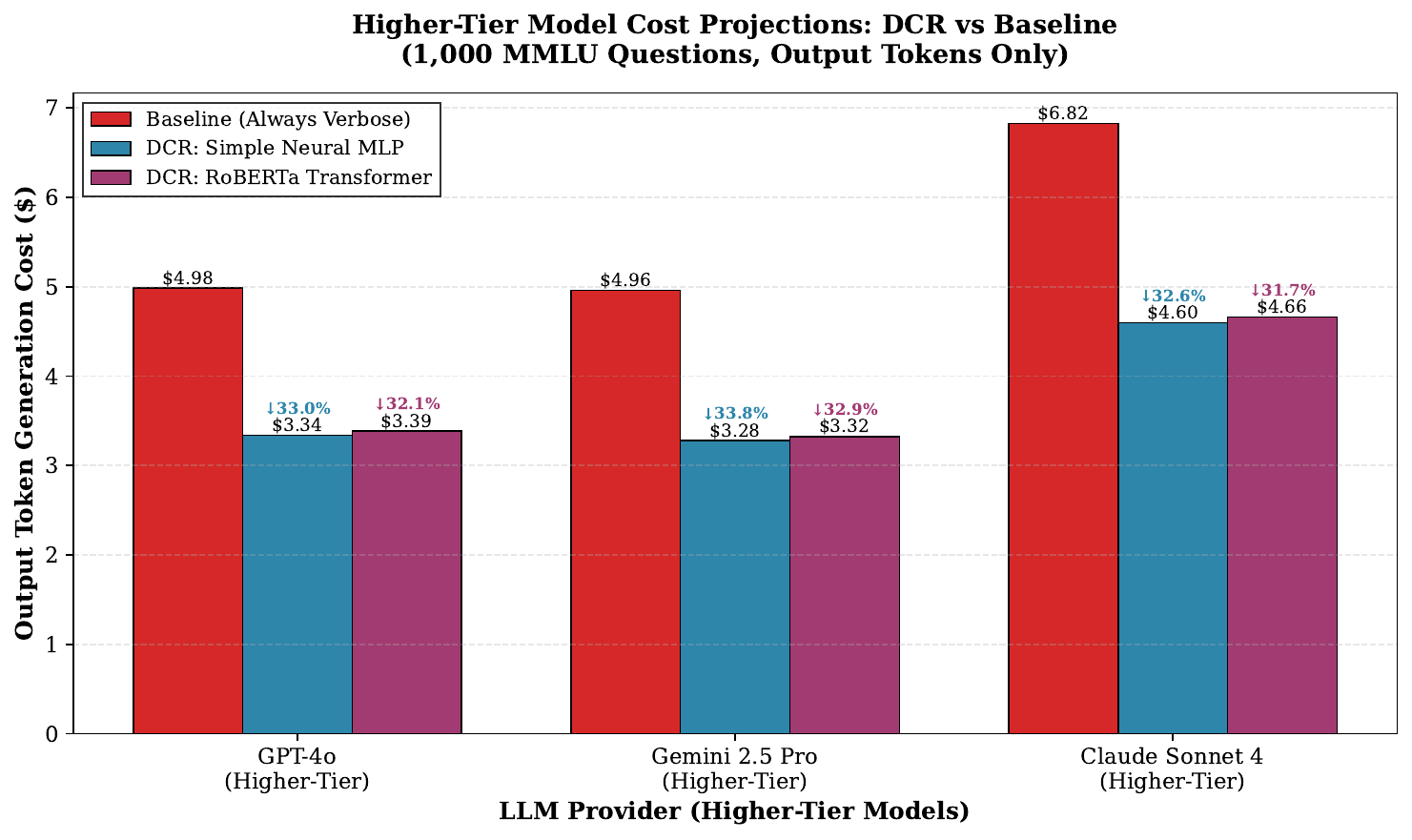}
    \caption{Cost Comparison: DTS vs Always-Verbose Baseline. Output token costs for 1,000 MMLU questions across higher-tier LLM models. DTS achieves substantial cost savings: \$1,646 (OpenAI GPT-4o), \$1,678 (Gemini 2.5 Pro), and \$2,225 (Claude Sonnet 4) per 1 million queries (multiply shown costs by 1,000). Cost savings scale with output token pricing multipliers (4-8×) and provider-specific token generation patterns.}
    \label{fig:cost_comparison}
\end{figure}

\subsection{Generalization to Other LLM Systems}

The principles underlying our approach extend naturally to any LLM-based system exhibiting the following characteristics:
\begin{itemize}
    \item Availability of multiple response generation strategies with varying computational or economic costs
    \item Capability to make routing decisions based on observable query characteristics
    \item Requirements for cost optimization while maintaining output quality standards
\end{itemize}

Practical applications span numerous domains. Retrieval-augmented generation systems could modulate retrieval depth based on query complexity. Code generation platforms might adjust test coverage levels according to task criticality. Customer service applications could calibrate response detail to match inquiry sophistication. Each context presents opportunities for intelligent resource allocation through query-aware routing.

\section{Limitations and Future Work}

\subsection{Current Limitations}

Several constraints merit acknowledgment in the current implementation. The MLP router's dependence on OpenAI's text-embedding-3-small model introduces vendor lock-in concerns. In the future, we could add support for other embedding services like Cohere or Sentence-BERT, cache embeddings to cut down on API calls, or build versions that don't depend on any specific model.

The dataset's focus on academic domains potentially limits insights into system performance on creative writing tasks, specialized professional contexts (medical or legal), or production query distributions encountered in deployed systems. Evaluation across more diverse benchmarks would strengthen generalizability claims.

Our template design process currently requires manual specification and tuning. We could also have the system learn what makes a good template by analyzing lots of real conversations. Plus, letting the system adjust templates based on what users like could make it work better with less manual tweaking.

\subsection{Future Research Directions}

Online learning would let the router keep improving from real usage data and user feedback, getting better over time instead of staying static after training.

We could also expand this to route between different LLMs entirely, not just templates. Imagine sending simple questions to cheaper models and complex ones to more powerful (and expensive) models—that could really optimize the cost-performance balance.

Additionally, investigating DTS applicability beyond English-language contexts presents important challenges. Cross-linguistic evaluation would test whether template selection principles generalize across languages and cultural contexts, potentially requiring language-specific adaptations or universal routing strategies.

\section{Conclusion}

Dynamic Template Selection addresses a fundamental inefficiency in contemporary LLM deployment: the pervasive application of verbose prompting strategies to queries of varying complexity. Through intelligent routing of queries to appropriately-sized response templates, DTS achieves substantial reductions in output token generation while maintaining response quality and completeness.

Our comprehensive evaluation, conducted on \numTestQuestions{} MMLU questions, provides robust empirical validation of the approach. Routing accuracy was assessed on held-out test data, with subsequent application of routing decisions across \numProviders{} major LLM providers (OpenAI GPT-4, Google Gemini 2.5 Pro, and Anthropic Claude Sonnet). Through \totalAPICalls{} production API calls, we measured actual token consumption patterns and cost implications. Both evaluated architectures---the lightweight MLP and fine-tuned RoBERTa transformer---demonstrate strong performance, achieving over 90\% routing accuracy with only 1.0 percentage point separation. Notably, the architecturally simpler MLP marginally outperforms the transformer model (90.5\% vs 89.5\%), challenging assumptions about the necessity of complex architectures for this task.

Template selection exhibits provider-agnostic properties, with routing accuracy remaining constant at 90.5\% across all tested platforms. Token reduction varies from 32.6\% (Claude) to 33.9\% (Gemini) based on each provider's generation patterns. Since output tokens cost 4--8$\times$ more than input tokens, the average \avgSimpleNeuralSavings{} reduction directly targets the most expensive part of API costs.

Implementation specifications enable straightforward reproduction and deployment in production environments.

A few important takeaways from our work: First, choose your architecture based on what you can actually deploy, not tiny performance differences. Second, saving output tokens matters way more than input tokens because of how providers price things. Third, the magnitude of achievable savings correlates with provider verbosity---platforms with naturally expansive generation patterns benefit most from adaptive routing.

For production deployments, DTS offers a practical pathway to substantial cost reduction without sacrificing response quality. The approach maintains 100\% API success rates while delivering meaningful token savings across diverse query types and provider platforms.

\bibliographystyle{plain}
\bibliography{references}

@article{wei2022chain,
  title={Chain-of-thought prompting elicits reasoning in large language models},
  author={Wei, Jason and others},
  journal={NeurIPS},
  year={2022}
}

@article{brown2020language,
  title={Language models are few-shot learners},
  author={Brown, Tom and others},
  journal={NeurIPS},
  year={2020}
}

@article{kojima2022large,
  title={Large language models are zero-shot reasoners},
  author={Kojima, Takeshi and others},
  journal={NeurIPS},
  year={2022}
}

@article{wang2022self,
  title={Self-consistency improves chain of thought reasoning in language models},
  author={Wang, Xuezhi and others},
  journal={ICLR},
  year={2022}
}

@article{yao2023tree,
  title={Tree of thoughts: Deliberate problem solving with large language models},
  author={Yao, Shunyu and others},
  journal={NeurIPS},
  year={2023}
}

@article{besta2024graph,
  title={Graph of thoughts: Solving elaborate problems with large language models},
  author={Besta, Maciej and others},
  journal={AAAI},
  year={2024}
}

@article{madaan2023self,
  title={Self-refine: Iterative refinement with self-feedback},
  author={Madaan, Aman and others},
  journal={NeurIPS},
  year={2023}
}

@article{sanh2019distilbert,
  title={DistilBERT, a distilled version of BERT},
  author={Sanh, Victor and others},
  journal={arXiv preprint},
  year={2019}
}

@article{hinton2015distilling,
  title={Distilling the knowledge in a neural network},
  author={Hinton, Geoffrey and others},
  journal={NIPS Deep Learning Workshop},
  year={2015}
}

@article{hendrycks2020measuring,
  title={Measuring massive multitask language understanding},
  author={Hendrycks, Dan and Burns, Collin and Basart, Steven and Zou, Andy and Mazeika, Mantas and Song, Dawn and Steinhardt, Jacob},
  journal={Proceedings of the International Conference on Learning Representations (ICLR)},
  year={2021}
}

@article{jiang2023llmlingua,
  title={LLMLingua: Compressing prompts for accelerated inference of large language models},
  author={Jiang, Huiqiang and others},
  journal={EMNLP},
  year={2023}
}

@article{zhou2023large,
  title={Large language models are human-level prompt engineers},
  author={Zhou, Yongchao and others},
  journal={ICLR},
  year={2023}
}

@article{lester2021power,
  title={The power of scale for parameter-efficient prompt tuning},
  author={Lester, Brian and others},
  journal={EMNLP},
  year={2021}
}

@article{wei2022finetuned,
  title={Finetuned language models are zero-shot learners},
  author={Wei, Jason and others},
  journal={ICLR},
  year={2022}
}

@article{rubin2022learning,
  title={Learning to retrieve prompts for in-context learning},
  author={Rubin, Ohad and others},
  journal={NAACL},
  year={2022}
}

@article{perez2021true,
  title={True few-shot learning with language models},
  author={Perez, Ethan and others},
  journal={NeurIPS},
  year={2021}
}

@article{reppert2023iterated,
  title={Iterated decomposition: Improving science Q\&A by supervising reasoning processes},
  author={Reppert, Justin and others},
  journal={arXiv preprint},
  year={2023}
}

@article{liu2022makes,
  title={What makes good in-context examples for GPT-3?},
  author={Liu, Jiachang and others},
  journal={ACL},
  year={2022}
}

@article{tishby2000information,
  title={The information bottleneck method},
  author={Tishby, Naftali and others},
  journal={arXiv preprint},
  year={2000}
}

@article{lewis2020retrieval,
  title={Retrieval-augmented generation for knowledge-intensive NLP tasks},
  author={Lewis, Patrick and others},
  journal={NeurIPS},
  year={2020}
}

\end{document}